\documentclass{article}

\PassOptionsToPackage{numbers, compress}{natbib}

\usepackage[abs]{overpic} 
\usepackage{pgfplots} 
\usepackage[preprint]{neurips_2022}

\usepackage[utf8]{inputenc} 
\usepackage[T1]{fontenc}    
\usepackage{hyperref}       
\hypersetup{hidelinks}
\usepackage{url}            
\usepackage{booktabs}       
\usepackage{amsfonts}       
\usepackage{nicefrac}       
\usepackage{microtype}      
\usepackage{xcolor}         
\usepackage{multirow}
\usepackage{wrapfig,lipsum,booktabs}

\usepackage{pgfplots} 
\usepackage[caption=false]{subfig}

\usepackage{pict2e} 
\usepackage{sansmath}
\usepackage{booktabs}
\usepackage{multirow}
\usepackage{float}

\usepackage{calc} 

\usepackage{color}
\newcommand{\chris}[1]{\textcolor{ForestGreen}{#1}}
\renewcommand{\chris}[1]{\textcolor[rgb]{0,0.0,0.0}{#1}}
\newcommand{\thms}[1]{\textcolor[rgb]{1,0.4,0.2}{#1}}
\renewcommand{\thms}[1]{\textcolor[rgb]{0,0.0,0.0}{#1}}
\newcommand{\semere}[1]{\textcolor[rgb]{1,0.2,0.4}{#1}}
\renewcommand{\semere}[1]{\textcolor[rgb]{0,0.0,0.0}{#1}}
\newcommand{\lucas}[1]{\textcolor[rgb]{0.5,0.5,0.7}{#1}}
\renewcommand{\lucas}[1]{\textcolor[rgb]{0,0.0,0.0}{#1}}
\newcommand{\newtext}[1]{\textcolor{NavyBlue}{#1}}
\renewcommand{\newtext}[1]{\textcolor[rgb]{0,0.0,0.0}{#1}}
\newcommand\dsim{{\sc{d-sim}}}
\newcommand\qsim{{\sc{q-sim}}}
\newcommand\dqsim{{\sc{dq-sim}}}
\newcommand\dqsimscore{{\sc{dq-sim}}-score}
\newcommand\dqsimrank{{\sc{dq-sim}}-rank}
\newcommand\CLS{{\sc{[cls]}}}
\newcommand\SEP{{\sc{[sep]}}}
\newcommand\mcq{{\sc{MCQ}}}

\newcommand{\delete}[1]{}
\newcommand{\rebuttal}[1]{\textcolor[rgb]{1,0.4,0.2}{#1}}
\renewcommand{\rebuttal}[1]{\textcolor[rgb]{0,0.0,0.0}{#1}}
\newcommand{\std}[1]{\tiny$\pm${#1}}

\newcommand{\etc}{etc.\ }
\newcommand{\eg}{e.g., }

\newcommand{\ie}{i.e., }
\newcommand{\vs}{vs.\ }
\newcommand{\etal}{\textit{et al.}\ }
\newcommand{\figref}[1]{Fig.~\ref{#1}}    
\newcommand{\Figref}[1]{Figure~\ref{#1}}  
\newcommand{\tabref}[1]{Table~\ref{#1}}
\newcommand{\Tabref}[1]{Table~\ref{#1}}
\newcommand{\secref}[1]{Section~\ref{#1}}

\newcommand{\appref}[1]{Appendix~\ref{#1}} 

\usepackage{amsmath} 
\usepackage[inline,shortlabels]{enumitem}

\usepackage[colorinlistoftodos,prependcaption,textsize=normalsize]{todonotes}

\usepackage{graphicx}
\graphicspath{ {./figures/} }

\usepackage[mathscr]{euscript}

\DeclareSymbolFont{rsfs}{U}{rsfs}{m}{n}
\DeclareSymbolFontAlphabet{\mathscrsfs}{rsfs}

\usepackage{amssymb}

\usepackage{multirow}
\usepackage{enumitem}

\title{Learning to Reuse Distractors to support \\Multiple
Choice Question Generation in Education}

\author{%
Semere Kiros ~Bitew$^{1}$\thanks{Corresponding author} \quad Amir Hadifar$^{1}$ \quad Lucas Sterckx\thanks{Lucas Sterckx, who is currently at LynxCare, contributed to this work while working in the T2K team, Ghent University-imec.} \quad Johannes Deleu$^1$ \quad Chris Develder$^1$ \\
\textbf{Thomas Demeester}$^1$ \\
$^1$IDLab, Ghent University – imec, \\
Technologiepark Zwijnaarde 126, 9052 Ghent, Belgium\\
\texttt{\{semerekiros.bitew,firstname.lastname\}@ugent.be}\\
\texttt{\{lucassterckx\}@gmail.com}\\
}

\begin{document}

\maketitle

\begin{abstract}
 Multiple choice questions (MCQs) are widely used in digital learning systems, as they allow for automating the assessment process. However, due to the increased digital literacy of students and the advent of social media platforms, MCQ
tests
are widely shared online, and teachers are continuously challenged to create new questions, which is an expensive and time-consuming task. 
A particularly sensitive aspect of
MCQ creation is 
to devise
relevant distractors, i.e., wrong answers that are not easily identifiable as being wrong. This paper studies how a large existing set of manually created answers and distractors for questions over a variety of domains, subjects, and languages can be leveraged  to help teachers 
in creating new MCQs, by the smart reuse of existing distractors. 
We built several data-driven models based on context-aware question and distractor representations, and compared them with static feature-based models. The proposed models are evaluated with automated metrics and in a realistic user test with teachers. Both automatic and human evaluations indicate that context-aware models consistently outperform a static feature-based approach. For our best-performing context-aware model, on average 3 distractors out of the 10 shown to teachers were rated as high-quality distractors. We 
create a performance benchmark, and make it public, to enable comparison between different approaches and to introduce a more standardized evaluation of the task. 
The benchmark contains a test of 298 educational questions covering multiple subjects \& languages and a 77k multilingual pool of distractor vocabulary for future research.

\end{abstract}

\section{Introduction}
\label{sec:intro}

Online learning has become an indispensable part of educational institutions. It has emerged as a necessary resource for students and schools all over the globe. The recent COVID-19 pandemic has made the transition to online learning even more pressing. 
One very important aspect of online learning is the need to generate homework, test, and exam exercises 
to aid and evaluate the learning progress of students~\cite{dunlosky2013improving}. Multiple choice questions (MCQs) are the most common form of exercises~\cite{gierl2017developing} in online education as they can easily be scored automatically. 
However, the construction of MCQs is time consuming \cite{davis2009tools} and there is a need to continuously generate new (variants of) questions, 
especially for testing, since students tend to share questions and correct answers from MCQs online (\eg through social media).

The rapid digitization of educational resources opens up opportunities to adopt artificial intelligence (AI) to automate the process of MCQ construction. A substantial number of questions already exist in a digital format, thus providing the required data as a first 
step toward building AI systems. The automation of MCQ construction could support both teachers and learners. Teachers could benefit from an increased efficiency in creating questions, in their already high workload. Students' learning experience could improve due to increased practice opportunities based on automatically generated exercises, \newtext{and if these systems are sufficiently accurate, they could power personalized learning~\cite{ma2014intelligent}}. 

A crucial step in MCQ creation is the generation of distractors~\cite{liu2017automatic}. Distractors are incorrect options that are related to the answer to some degree. The quality of an MCQ heavily depends on the quality of distractors \cite{davis2009tools}. If the distractors do not sufficiently challenge learners%
, picking the correct answer becomes easy, ultimately degrading the discriminative power of the question. The automatic suggestion of distractors will be the focus of this paper.

Several works have already proposed distractor generation techniques for automatic MCQ creation, mostly based on selecting distractors according to their similarity to the correct answer. 
In general, two approaches are used to measure the similarity between distractors and an answer: graph-based and corpus-based methods. \emph{Graph-based} approaches use the semantic distance between concepts in the graph as a similarity measure. In language learning applications, typically 
WordNet~\cite{mitkov2009semantic, pino2008selection} is used to generate distractors, 
while for factoid questions domain-specific (ontologies) are used to generate distractors~\cite{papasalouros2008automatic, faizan2018automatic, leo2019ontology, alsubait2014generating}. 
In \emph{corpus-based methods}, similarity between distractors and answers has been defined as having similar frequency count~\cite{coniam1997preliminary}, belonging to the same POS class~\cite{goto2010automatic}, having a high co-occurrence likelihood~\cite{hill2016automatic}, having similar phonetic \chris{and} 
morphological features~\cite{pino2008selection}, 
and being nearby in embedding spaces~\cite{kumar2015automatic,guo2016questimator,jiang2017distractor}. Other works such as~\cite{liu2017automatic, liang2017distractor,liang2018distractor,liu2016automatic} 
use machine learning models to generate distractors by using a combination of the previous features and other types of information such as tf-idf scores. 


While the current state-of-the-art in MCQ creation is promising, we see \thms{a number of limitations. First of all, existing models are often \emph{domain specific}.} 
Indeed, the proposed techniques are tailored to the application and distractor types. In language learning, such as vocabulary, grammar or tense usage exercises, typically similarity based on basic syntactic \chris{and} 
statistical information works well: frequency, POS information\chris{,} \etc In other domains, such as science, health, history, geography, etc., distractors should be selected on deeper understanding of context and semantics, and the current methods fail to capture such information. 


The second limitation, \emph{language dependency}, is especially applicable to factoids. \thms{Models should be agnostic to language because facts do not change with languages. Moreover, building a new model for each language could be daunting task as it would require enough training data for each language.} 


In this work, we study how the automatic retrieval of distractors can facilitate the efficient construction of MCQs. We use a high-quality large dataset of question, answer, distractor triples that are diverse in terms of language, domain, and type of questions. \thms{Our dataset was made available by a commercial organization active in the field of e-assessment (see \secref{sec:data}), and is therefore representative for the educational domain, with a total of 62k MCQ, none of them identical, encompassing only 92k different answers and distractors. Despite an average of 2.4 distractors per question, there is a large reuse of distractors over different questions. This motivates our premise to retrieve and \emph{reuse} distractors for new questions.} We make use of the latest data-driven Natural Language Processing (NLP) techniques to retrieve candidate distractors. We propose \emph{context-aware multilingual models} that are based on deep neural network models that select distractors by taking into account the context of the question. They are also able to handle variety of distractors in terms of length and type. We compare our proposed models to a competitive \emph{feature-based} baseline that is based on classical machine learning methods trained on several handcrafted features.



The methods are evaluated for distractor quality using automated metrics and a real-world user test with teachers. Both the automatic evaluation and the user study with teachers 
indicate that the proposed context-aware methods outperform the feature-based baseline.  
Our contribution can be summarized as follows:
\begin{itemize}

   \item We built three multilingual Transformer-based distractor retrieval models that suggest distractors to teachers for multiple subjects in different languages. The first model (\secref{subsection:dqsim}) requires similar distractors to have similar semantic representations, while the second (\secref{subsection:qsim}) learns similar representations for similar questions, and the last combines the complementary advantages of 
  of these two models (\secref{subsection:dqsim}).
    \item \rebuttal{We performed a user study with teachers to evaluate the quality of distractors proposed by the models, based on a four-level annotation scheme designed for that purpose.}
    \item \rebuttal{The evaluation of our best model on in-distribution held-out data reveals an average increase of 20.4\% in terms of recall at 10, compared to our baseline model adapted from~\cite{liang2018distractor}. 
    The teacher-based annotations on language learning exercises
    show an increase by 4.3\% in the fraction of good distractors among the top 10 results, compared to teacher annotations for the same baseline. For factoid questions, the fraction of quality distractors more than doubles w.r.t.~the baseline, with an improvement of 15.3\%.}
    \item We \rebuttal{released\footnote{\url{https://dx.doi.org/10.21227/gnpy-d910} or \url{https://github.com/semerekiros/dist-retrieval}}} a test-set of educational questions of 6 subjects with 50 MCQs per subject and annotated distractors, and 77k size distractor vocabulary as benchmark to stimulate further research. The dataset, which is made by experts, contains multilingual and multi-domain 
    distractors. 
\end{itemize}

The remainder of the paper is organized as follows: \secref{sec:relatedworks} describes the relevant work in MCQs in general and distractor generation in particular. \secref{sec:methodology} introduces the dataset, explains the details of the proposed methods and the evaluation setup of the user study with teachers. In \secref{sec:resultsanddiscussion}, the results of both the user study and automated evaluations is reported. And finally, in \secref{sec:conclusionandfuturework}, we present the conclusion, lines for future work, and limitation\chris{s} of our proposed models.

\section{Related work}
\label{sec:relatedworks}

\subsection{MCQs in Education}
 
Multiple choice questions (MCQs) are widely used forms of exercises that require students to select the best possible answer from a set of given options. They are used in the context of learning, and assessing learners' knowledge \chris{and} 
skills.
MCQs are categorized as objective types of questions because they primarily deal with the facts or knowledge embedded in a text rather than subjective opinions~\cite{ch2018automatic}.  
It has been shown that recalling information in response to a multiple-choice test question bolsters
memorizing capability,  which leads to better retention of that information over time. It can also change the way information is represented in memory, potentially resulting in deeper understanding~\cite{butler2018multiple} of concepts.    

An MCQ item consists of three elements:
\begin{itemize}
    \setlength{\itemindent}{1em}
    \item \emph{stem}: is the question, statement, or lead-in to the question.
    \item \emph{key}: the correct answer.
    \item \emph{distractors}: alternative answers meant to challenge students' understanding of the topic. 
\end{itemize}
For example, consider the {\mcq} in the first row of \tabref{tab:annotationschemeexamples}: the stem of the \mcq{} is \emph{``Which inhabitants are not happy with Ethiopia's plans of the Nile?"}. Four potential answers 
are given with the question. Among these, the correct answer is \emph{``Egyptians"}, which is the key. The 
alternatives are the distractors.

\newtext{{\mcq}s are used in several teaching domains such as information technology~\cite{woodford2004using}, health~\cite{brady2005assessment,collins2006education}, historical knowledge~\cite{10.2307/40543353}, \etc They are also commonly used in standardized tests such as GRE and TOEFL. {\mcq}s are preferred to other question formats because } 
they are easy to score, and students 
\chris{can} also answer them relatively quickly since typing responses is not required. \newtext{Moreover, {\mcq}s enable a high level of test validity if they are drawn from a representative sample of the content areas that make up the pre-determined learning outcomes~\cite{collins2006education}.}
The most time-consuming and non-trivial task in constructing \mcq{} is distractor generation~\cite{davis2009tools,liang2018distractor}. Distractors should be plausible enough to force learners to put some thought before selecting the correct answer.
Preparing good multiple-choice questions is a skill that requires formal training~\cite{abdulghani2015faculty, naeem2012faculty}.
Moreover, several \mcq{} item writing guidelines are used by content specialists when they prepare educational tests.
These guidelines also include recommendations for developing and using distractors~\cite{haladyna1989taxonomy, haladyna2013developing,moreno2015guidelines}.
Despite these guidelines, inexperienced teachers may still construct poor MCQs due to lack of training and limited time~\cite{vyas2008multiple}. 

Besides reducing teachers' workloads, the automation of the distractor generation could potentially correct some minor mistakes made by teachers.
For example, one of the rules suggested by~\cite{haladyna1989taxonomy} says:  ``the length of distractors and the key should be about the same''. 
Such property could be easily integrated in the automation process.

MCQs \chris{also} have drawbacks; they are typically used to measure lower-order levels of knowledge, and guesswork can be a factor in answering a question with a limited number of alternatives.
\newtext{Furthermore, because of a few missing details, learners' partial understanding of a topic may not be sufficient to correctly answer a question, 
resulting in partial knowledge not being credited by {\mcq}s~\cite{butler2018multiple}.}
Nonetheless, \mcq{}s are still extensively utilized in large-scale tests since they are efficient to administer and easy to score objectively~\cite{gierl2017developing}.
   
\subsection{Distractor Generation} 

Many strategies have been developed for generating distractors for a given question. The most common approach is to select a distractor based on its similarity to the key for a given question. Many researchers approximate the similarity between distractor and key according to WordNet~\cite{mitkov2006computer,mitkov2003computer,lin2007automatic}. WordNet~\cite{miller1990introduction} is a lexical database that groups words into sets 
of synonyms, and concepts semantically close to the key are used as distractors. The usage of such lexical databases is sound for language or vocabulary learning but not for factoid type questions. We instead provide a more general approach that could be used for both tasks, and instead of only using the key as the source of information while suggesting distractors, we also make use of the stem. 

For learning factual knowledge, several works rely on the use of specific domain ontology as a proxy for similarity.
Papasalouros~\etal \cite{papasalouros2008automatic} employ several ontology-based strategies to generate distractors for MCQ questionnaires.
For example, they generate ``Brussels is a mountain" as a good distractor for an answer ``Everest is a mountain" because both concept \emph{City} and concept \emph{Mountain} share the parent 
concept \emph{Location}.
Another very similar work by Lopetegui \etal \cite{lopetegui2015novel} selects distractors that are declared siblings of the answer in 
\chris{a} domain\chris{-}specific ontology.
The work by Leo~\etal \cite{leo2019ontology} improves \chris{upon} the previous works by generating multi-word distractors from an ontology in the medical domain.
Other works that rely on knowledge bases 
\chris{apply}
query relaxation methods\chris{,} where the queries used to generate the keys were slightly relaxed to generate distractors that share similar features with the key  \cite{seyler2017knowledge,faizan2018automatic,stasaski2017multiple}. While the methods in these works are dependent on 
their respective ontologies, we provide an approach that is ontology-agnostic and instead uses 
contextual similarity between distractors and questions.   

Another 
\chris{line of works} for distractor generation 
\chris{uses} machine-learning models.
Liu~\etal \cite{liu2017automatic} use a regression model based on characteristics such as character glyph, phonological, and semantic similarity for generating distractors in Chinese. Liang~\etal \cite{liang2018distractor} use two methods to rank distractors in the domain of school sciences.
\chris{The first method adopts} machine learning classifiers on manually engineered features (\ie edit distance, POS similarity, etc.) to rank distractors.
\chris{The second uses} generative adversarial networks to rank distractors.
Our baseline method is inspired by their first approach but was made to account for the multilingual nature of our dataset by extending the feature set. 

There have also been a number of works on 
generating distractors in the context of machine comprehension~\cite{lai2017race}.
Distractor generation strategies that fall in this category assume access to a contextual resource such as a book chapter, an article or a wikipedia page where the MCQ was produced from.
The aim is then to generate a distractor that takes into account the reading comprehension text, and a pair \chris{composed} of \chris{the} question and its correct answer that originated from the text~\cite{gao2019generating,zhu2018hierarchical,chung2020bert}.
This line of work is incomparable to our work because we do not have access to an external contextual resource 
the questions were prepared from. 

In this paper, we focus on building one model that is able to suggest candidate distractors for teachers both in the context of language and factual knowledge learning.
Unlike previous methods, we tackle distractor generation with a multilingual dataset.
Our distractors are diverse both in terms of domain and language.
Moreover, the distractors are not limited to single words only. 


\section{Methodology}
\label{sec:methodology}

In this section, we formally define distractor generation as a ranking problem; describe our datasets; describe in detail the feature-based baseline and proposed context-aware models including their training strategies \& prediction mechanisms.

\subsection{Task Definition: Distractor Retrieval}
\label{sec:taskdefinition}

We assume access to a 
distractor candidate set $\mathcal{D}$ and a training MCQ dataset $\mathcal{M}$. \thms{Note that $\mathcal{D}$ can be obtained by pooling all answers (key\chris{s} and distractors) from $\mathcal{M}$ (as in our experimental setting), but could also be augmented, for example, with keywords extracted from particular source texts. We formally write
\chris{$\mathcal{M} = \left \{ (s_{i}, k_{i}, \mathcal{D}_i) | i=1, \ldots, N\right\}$.}
where for each item $i$ among all $N$ available MCQs, $s_{i}$ refers to the question stem, $k_{i}$ is the correct answer key, and $\mathcal{D}_{i} = \big\{d_{i}^{(1)},...,d_{i}^{(m_i)} \big\} \subseteq \mathcal{D}$ are the distractors in the MCQ linked to $s_{i}$ and $k_{i}$. }
The aim of the distractor retrieval task is to learn a point-wise ranking score $r_i(d): (s_{i}, k_{i}, d) \to \left [  0, 1\right ]$ for all $d \in \mathcal{D}$, such that distractors in $\mathcal{D}_{i}$ are ranked higher than those in $\mathcal{D}\setminus \mathcal{D}_{i}$, when sorted according to the decreasing score $r_i(d)$. \\
This task definition resembles the metric learning \cite{kulis2013metric} problem in information retrieval. To learn the ranking function, we propose two types of models: feature-based models and context-aware \chris{n}eural \chris{n}etworks. 

\begin{table}[t!]
\centering
\caption{The statistics of our dataset}
\label{tab:dataset}
\begin{tabular}{llll}
\hline
 & \textbf{Train} & \textbf{Validation} & \textbf{Test} \\ \hline
\# Questions & 61758 & 600 & 500 \\
\# Distractors per question & 2.4 & 2.3 & 2.3 \\
Avg question length & 27.8 tokens & 28.1 tokens & 27.6 tokens \\
Avg distractor length & 2.2 tokens & 2.3 tokens & 2.1 tokens \\
Avg answer length & 2.2 tokens & 2.3 tokens & 2.2 tokens \\ \hline
Total \# distractors & 94,205 & - & - \\ 
Total \# distractors $\leq$ 6 tokens & 77,505 & - & - \\\hline
\end{tabular}

\end{table}

\subsection{Data}
\label{sec:data}

In this section, we describe our datasets, namely: \begin{enumerate*}[(i)]
    \item \emph{Televic dataset}, a big dataset 
    \chris{that} we used to train our models.
    \item \emph{Wezooz dataset}, a small-scale external test set used for evaluation.
\end{enumerate*}

\subsubsection{Televic dataset}
\label{sec:televicdataset}
This data is gathered through Televic Education's platform 
assessmentQ.\footnote{\url{https://www.televic-education.com/en/assessmentq}} The tool is a comprehensive online platform for interactive workforce learning and high-stakes exams. It allows teachers to compose their questions and answers for practice and assessment. As a result, the dataset is made up of a large and high-quality set of questions, answers and distractors, 
manually created 
by experts in their respective fields. It encompasses a wide range of domains, subjects, and languages, without however any metadata on the particular course subjects that apply to the individual items.

We randomly divide our dataset into train/validation/test splits. 
We discard distractors with more than 6 tokens \lucas{as they are very rare} and unlikely to be reused in different contexts. We keep questions with at least one distractor. \Tabref{tab:dataset} summarizes the statistics of our dataset. 
The dataset contains around 62k MCQs in total. The size of the dataset is relatively large when compared to previously reported educational MCQ datasets such as SCiQ \cite{welbl2017crowdsourcing}, and MCQL \cite{liang2018distractor} which contain 13.7K and 7.1K MCQs respectively.  On average, a question has more than 2 distractors, \rebuttal{and contains exactly one answer}. We use all the answer keys and distractors in the preprocessed dataset as the pool of candidate distractors (\ie list of 77,505 filtered distractors) for proposing distractors for any new question. 

\begin{figure}[t]
\subfloat[]{%
\label{fig:answer_length}
\resizebox{0.5\textwidth}{!}{%
 \begin{tikzpicture} 
\begin{axis} [
    axis x line*=bottom,
    axis y line*=left,
    ybar,
    ymin=0,
    xlabel={No of words per distractors},
    ylabel={Frequency \%}, ]

\addplot coordinates {(1, 32.76389301059314) (4, 11.02150884481891) (5, 9.901552197979433) (2, 26.134472213978814) (3, 16.994180870417917) (6, 3.184392862211785)};
\end{axis}
 
\end{tikzpicture} %
}%
}
\subfloat[]{%
  \label{fig:language_distribution}
  \resizebox{0.5\textwidth}{!}{%
 \begin{tikzpicture}
\begin{axis} [
x tick label style={
		/pgf/number format/1000 sep=},
    axis x line*=bottom,
    axis y line*=left,
    ybar interval=0.5,
    xmajorgrids=false,
    ylabel={Frequency \%},
    xlabel={Language\footnotemark},
    ymin=0,
    symbolic x coords={nl,fr,en,de, es,it,zh,pt,af,la},
    x tick label style={anchor=north, text height=1.5ex, align=right}
    ]
\addplot coordinates {(nl, 56.271278118934745)
(fr, 28.688472429921408)
(en, 9.779184829297783)
(de, 1.8836106780362085)
(es, 1.1088485157020587)
(it, 0.7047631168665882)
(zh, 0.5488561519615641)
(pt, 0.194088262432785)
(af, 0.1702249514779344)
(la, 0.14636164052308379)
};
\end{axis}
\end{tikzpicture} %
}%
}

\caption{(a) distractor length in number of tokens and (b) language distribution for the Televic dataset.}

\end{figure}
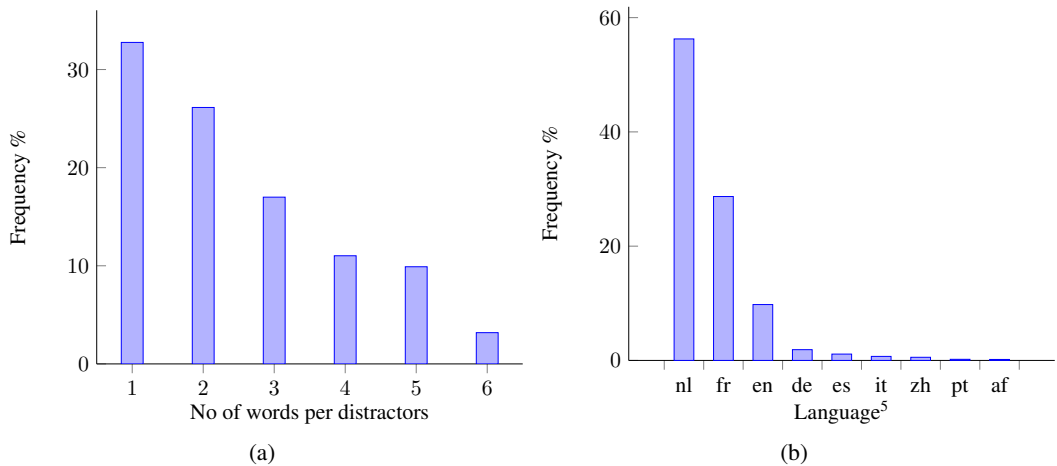

The distractors in the dataset are not limited to single word distractors. More than 65\% of the distractors contain two or more words as can be seen in \figref{fig:answer_length}.

\footnotetext{We used ISO 639-1:2002 standard for names of languages.}

The data stems from multiple languages. \Figref{fig:language_distribution} shows the language distribution as detected by an off-the-shelf language classifier.\footnote{We used the \emph{langid} language classifier:  \url{https://github.com/saffsd/langid.py}} 
Given that Televic is a Belgian company, more than 50\% of the questions are in Dutch, while French and English are the next most 
common languages in the dataset.

Another dimension of the dataset is its domain diversity.
It comprises questions about language/vocabulary learning (\eg French and English) and factoids \lucas{covering} subjects such as 
Math, Health, History, Geography, and Sciences.
Besides material from secondary school education, it covers materials from assessment tasks for professionals such as training in hospitals or manufacturing firms.
The data is anonymized and contains no customer information.

Depending on the question type we observe different types of distractors. \begin{enumerate*}[(1)]
    \item Factoid distractors: 
    names of people, locations, organizations, concepts, dates.
    \item Distractors with numerical elements, such as multiples, factors, rounding errors, \etc
    \item Language distractors: spelling, grammatical, tense, \etc 
\end{enumerate*}
However, the proposed models are agnostic of the type and origin of the data, and the automated evaluation on the Televic test set contains a random sample covering the different question types and origins (see \secref{sec:automaticevaluation}).
\rebuttal{
Note that although our dataset is a real-world commercial dataset, it only contains single-answer \mcq{}s. However, the models we will put forward, could be readily extended towards multiple-answer \mcq{}s, if such data were available.}

\subsubsection{WeZooz dataset}
\label{sec:wezoozdataset}
This data is a small-scale test set of questions gathered from WeZooz Academy\chris{,}\footnote{\url{https://www.wezoozacademy.be/}} 
\chris{which} is a Flanders-based company providing an online platform with digital teaching materials for secondary school students and teachers.
We selected four subjects; Natural sciences, Geography, Biology and History.
Each subject was made to contain a fixed list of 50 questions that were randomly selected, \thms{and augmented with distractor annotations by teachers for these respective subjects (see \secref{sec:experimentaldesign}).  
Note that this is an \emph{external} test set, in the sense that the data distribution in the training set is not necessarily representative for this test set. This serves as a proof-of-concept for the general validity of our proposed method and models to specific use cases. 
}


\subsection{Feature-based \thms{Distractor Scoring}}

We built a strong feature-based model as our baseline. Feature-based models are a class of machine learning models that require a pre-specified set of handcrafted features as input. We designed 20 types of features capturing similarity between questions, answers, and the collection of candidate distractors. Formally, given a triplet \chris{$(s,k, d)$} of question stem, key and distractor, 
our feature-based model first maps the input into a 20-dimensional feature vector $\phi\left(s, k, d\right)\;\chris{\in}\;\mathbb{R}^{20}$, after which a classifier is trained to score the triplets according to compatibility of the question-answer-distractor combination. Our set of features can be segmented into four categories which are described below. A more detailed explanation of each feature can be found in \appref{appendix:featureset}.

\begin{enumerate}[(i)] 
    \item \emph{Morphological Features}: this category contains features that are related to the form and shape of words that occur in our $(s, k, d)$ triplets. This includes features such as edit distance, difference in token length, longest common suffix between $k$ \& $d$, \chris{etc.}
    
    \item \emph{Static embedding based features}: We trained a Word2Vec model \cite{mikolov2013efficient} on our dataset to learn static embeddings for the distractors. We treat distractors and answers attached to the same question as chunks sharing similar context. The objective is to learn a vector space in which their representations will also be closer. We leverage the embedding representations to extract several numerical features. For example, we calculate the cosine similarity and word mover's distance \cite{kusner2015word} between the embeddings of $d$ \& $k$.
    
    \item \emph{Language Prior}: since our data is multilingual we also calculate the prior probability of the candidate distractor matching with the language of the question, and attach it to each feature vector.
    
    \item \emph{Corpus-based Features}: this category contains features that are derived from the statistics of words in the corpus. It includes features that such as the frequency of a distractor in the dataset and the inverse document frequency of distractors. 
\end{enumerate}
        
As classifier, we apply a \emph{logistic regression} model to distinguish feature representations of actual question-answer-distractor triplets, present in the training, from triplets for which the distractor components belong to different question-answer combinations, sampled randomly. During training, the model's parameters are set to output high scores for actual triplets while the model is penalized for predicting high scores for others.

\subsection{Context-aware Neural \thms{Distractor Scoring}}

\begin{figure*}[ht]

\centering
\includegraphics[width=\linewidth]{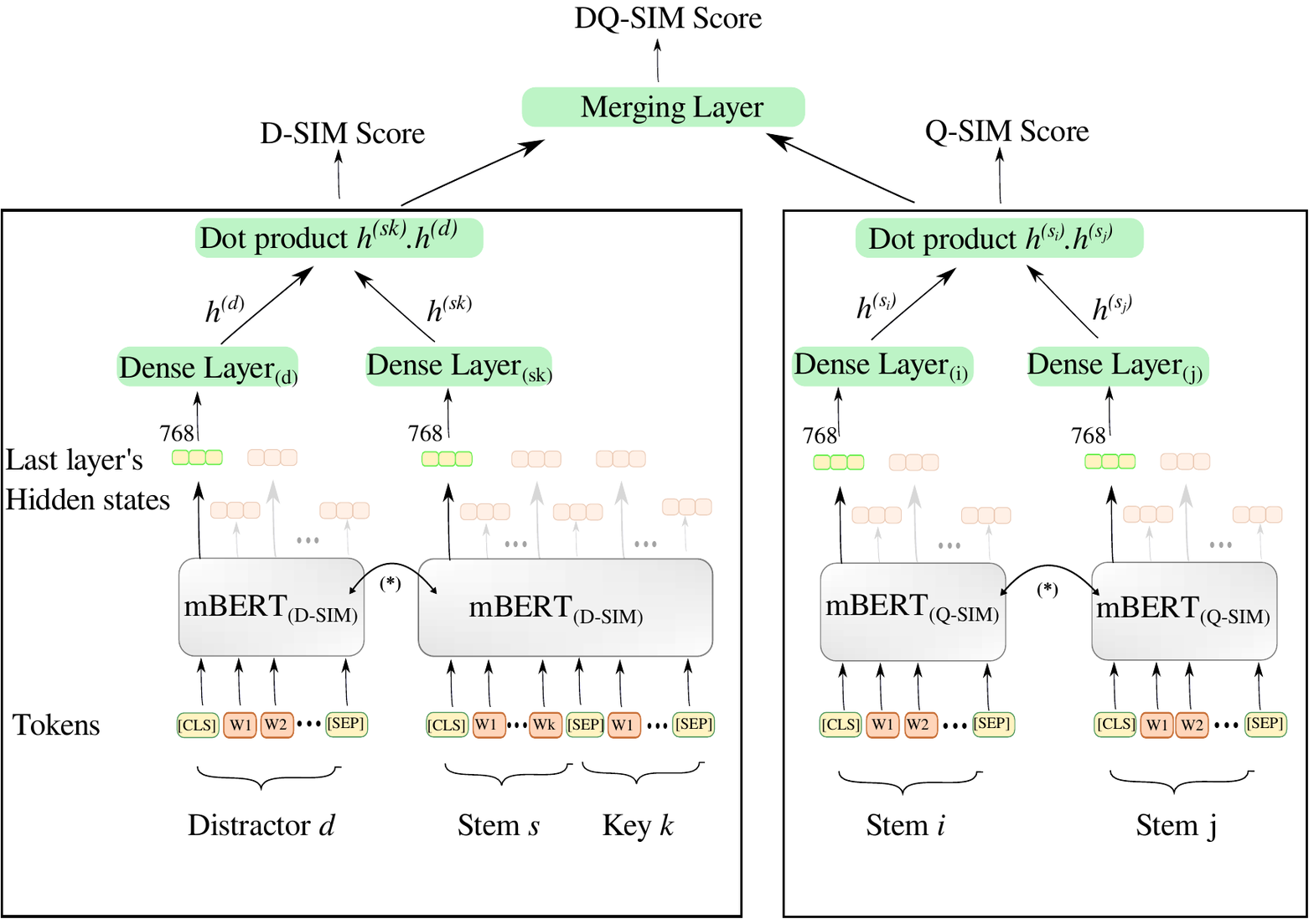}
\caption{Our proposed context-aware distractor retrieval systems. For the \dsim{} model (\ie left), distractor $d$ and concatenation of the stem $s$ \& key $k$ separated by \SEP{} are fed into the \underline{same} mBERT\textsubscript{(D-SIM)} encoder, and then their respective vector representations at \CLS{} are used as inputs to two \underline{different} dense layers that do not share parameters. The outputs of these dense layers, $h^{(d)}$ \& $h^{(sk)}$ are used to calculate the similarity between $d$ \& $s\SEP k$ using the dot product. Similarly for \qsim{} (\ie right), two question stems $i$ \& $j$ are encoded separately using the \underline{same} mBERT\textsubscript{(Q-SIM)}, and their respective \CLS output vectors are fed into two \underline{different} dense layers (\ie dense layer\textsubscript{(i)} \& dense layer\textsubscript{(j)}) to produce their corresponding representations $h^{(s_{i)}}$ \& $h^{(s_{j})}$. These are used to calculate their similarity between the two stems using dot product. The \dqsim{} model (\ie top) linearly combines the two models using a merging layer with an $\alpha$ parameter. $(\star)$ denotes parameter reuse by the encoders.}
\label{fig:mbert}
\end{figure*}

Advanced context-aware neural models, unlike traditional feature-based models, do not require manual feature engineering. They have the ability to represent words depending on their semantic role and context in the considered text. 
In this work, we primarily focus on such context-aware models called \emph{transformers} \cite{vaswani2017attention}, which provide rich representations, and proved to achieve state-of-the-art results for many tasks in NLP such as question answering~\cite{radford2018improving}, machine translation~\cite{edunov2018understanding}, and text summarization~\cite{zhong2020extractive}.  
A transformer is a deep neural network 
that uses a self-attention mechanism to assign importance weights to every part of the input sequence in how they are related to all other parts of the input. \thms{Transformers can scale to very large numbers of trainable parameters, stacked into very deep networks, which can still be trained very efficiently on parallel GPU hardware and thus learn from very large amounts of data.}
In NLP, such models are often trained on large unlabeled corpora to learn the inherent word and sentence level correlations (\ie to model language structure) between varying contexts. This process is called \emph{pretraining}, and downstream NLP tasks usually rely on \thms{such a pretrained generic model to be \emph{finetuned} to their more specific} needs instead of training a new model from scratch. \thms{Leveraging the knowledge gained during a generic pretraining process to improve prediction effectiveness for a specific supervised learning task, is a form of \emph{transfer learning}~\cite{pan2009survey,goodfellow2016deep}}. 
A common language task often used for pretraining transformer models called \emph{masked language modelling} (MLM) requires masking a portion of the input text and then training 
a model to predict the masked tokens --- in other words, to reconstruct the 
original non-masked input. BERT (
{Bidirectional Encoder Representations from transformers})~\cite{devlin2018bert} is the most popular pretrained masked language model and has been widely used in many downstream tasks such as question answering \& generation, machine reading comprehension, and machine translation, by fine-tuning it using a labelled dataset that provides supervision signal. 

In this work, we 
\thms{present models to rank and retrieve distractors, based on such a pretrained transformer text encoder, which we finetuned by}
%
requiring similar distractors to have similar representations, and similar questions also to have similar representation. \thms{In the following paragraphs, we provide a detailed description of these models, } 
visualized in \figref{fig:mbert}, followed by a description of the training procedure and the inference mechanism.

\subsubsection{Distractor similarity based model (\dsim)}
\label{subsection:dsim}
We hypothesize that distractors co-occurring within the same MCQ item are semantically related \thms{through their link with the corresponding question stem and answer key.}
\thms{Following that hypothesis, the {\dsim} model is designed (and trained) to yield a similar vector representation for a given (stem, key) pair $(s_i, k_i)$, as each of the corresponding distractors $d_i$. 
Following the same logic, all candidate distractors $d\in\mathcal{D}$ can then be scored in terms of their similarity (in representation space) with a given new (stem, key) pair, after which the top candidates are returned by the model as likely valid distractors.
}
%
\thms{We use the pretrained multilingual BERT (mBERT) encoder \cite{devlin2018bert}, \newtext{followed by a fully connected linear layer (\ie dense layer)} to obtain initial representations for a (stem, key) pair, as well as for the distractors. 
\newtext{We designed our model} in a bi-encoder setting} 
inspired by \cite{guo2020multireqa}, \thms{and schematically shown on the left-hand side of \figref{fig:mbert}}.
\thms{The distractor $d$} is fed into the mBERT 
encoder, and the output representation of the $[CLS]$ token
\footnote{[CLS] is a special token that is prepended to the input, and its corresponding output representation is pretrained to represent the entire sequence that is used for classification tasks.} 
\newtext{is used as an input to the dense layer. The output from the dense layer is taken as the corresponding representation $h_{d}$.} \delete{is taken as the corresponding representation $h_{d}$.} 
\thms{The considered stem and key are concatenated into a single sequence of tokens\footnote{The often used [SEP] token is a special token known by the model, that separates input sentences.} as ``$s_i\;[SEP]\;k_i$'', which}
is fed into the \emph{same} mBERT encoder \thms{(i.e., with parameter reuse, as 
indicated by the double arrow in \figref{fig:mbert}).} 
Similar to the distractor embedding, we take the \CLS{} token representation \newtext{and feed it to the dense layer (\ie \emph{different} dense layer with no parameter sharing), and take its output as the vector representation of the key-aware stem $h_{sk}.$} \delete{at the output side as the vector representation of the \thms{key-aware stem $h_{sk}$.}}
Finally, \thms{the similarity score between $(s_i, k_i)$ and $d$ is obtained as the dot product between their respective representations:}
\[r_i^{\text{\dsim}}(d) = h^{(sk)}_{i}\cdot h^{(d)}\]
During training, the encoder is fine-tuned to \thms{achieve higher scores for compatible stem/key and distractor combinations, and lower scores for incompatible ones (as described in \secref{sec:training} in more detail).}

\begin{table*}[t]
\centering
\scriptsize
\caption{{\qsim} training data examples.}
\label{tab:q-sim-example}

\begin{tabular}{l p{4cm}p{5cm}}
\toprule
\textbf{Distractor/Answer} & \textbf{Associated Questions}&             \textbf{Description}                                                                                   \\ \midrule
\multirow{6}{2cm}{koolhydraten en vetten}     & 1. Welke groepen voedingsstoffen leveren vooral energie?  & \multirow{2}{5cm} {Factoid questions with multi-word distractor in Dutch. }                                                                   \\
                           & 2. Welke voedselcomponenten kunnen stoffen leveren die zowel bij assimilatie als bij dissimilatie in cellen worden gebruikt?   
                           \\ \midrule
\multirow{2}{4em}{surrounded}  & 1. The guest house is \ldots\ on the countryside.  & \multirow{2}{5cm}{A fill in the gap question for English language learning.} \\
                           & 2.  The valley was \ldots\ by forests.                                                                                           \\ \midrule
\multirow{2}{4em}{Marokko}  & 1. Welk land is in 2011 gesplitst door het langdurig conflict in Darfur ?   & \multirow{2}{5cm}{A combination of fill-in the gap and normal questions}                                                 \\
                            & 2. Rabat is de hoofdstad van \ldots                                           \\ \bottomrule                                     
\end{tabular}
\end{table*}

\subsubsection{Question similarity based model (\qsim)}
\label{subsection:qsim}
This model is based on the assumption that \thms{different questions that share one or more distractors or answer keys 
are likely semantically related, such that}
their associated distractors could be used as good candidate distractors for one another. To accomplish this, we first rearrange the training data in such a way that these questions, sharing at least one distractor or key, are clustered together (see \tabref{tab:q-sim-example} for an example). Then, we train our \qsim{} model to produce similar representation for question stem pairs ($s_i, s_j$) that are in the same cluster. 
The right-hand side of~\figref{fig:mbert} depicts the \qsim{} model, again based on a bi-encoder architecture. 
\thms{The stem representation $h^{(s)}_i$ for a question \mcq$_i$ is again obtained through an mBERT encoder, \newtext{followed by a fully connected linear layer,} similarly to $h^{(sk)}_i$ but ignoring the question key.
The \qsim{} scoring function is defined as}
\[r_i^{\text{\qsim}}(d_{j}) = h_{i}^{(s)}\cdot h_{j}^{(s)}\]
\thms{and can be interpreted as follows. For a given question \mcq$_i$, its stem representation $h_i^{(s)}$ is compared through dot product similarity with the representation of any candidate distractor $d_j$ originating from a question \mcq$_j$. The particular representation of $d_j$ assumed in \qsim{} is in fact \mcq$_j$'s stem representation $h_j^{(s)}$.  
Note that \qsim{} does not allow making a distinction in terms of score between different distractors from the same \mcq. Candidate distractors with the same score are considered equally likely according to \qsim{}, and ranked in an arbitrary order. }%
\rebuttal{Based on the intuition outlined above, more complex formulations for \qsim{} can be designed, for example with a feature characterizing the nature of the pairwise comparison (i.e., the actual answers of the considered questions, two of their respective distractors, or the answer for the one and a distractor for the other). However, given the already significant improvement of the presented basic \qsim{} formulation (see \secref{sec:automaticevaluation}), we chose to include only that model in our evaluation. In fact, its simple intuitive formulation makes it straightforward to explain to teachers, which is an important aspect in their trust in the model~\cite{khosravi2022explainable}.}




\subsubsection{Distractor and Question similarity model (\dqsim)}
\label{subsection:dqsim}
This model combines the previous two models using a merging layer \thms{(visualized on top of Fig.~\ref{fig:mbert})}, \rebuttal{based on the intuition that a well-chosen combined model may benefit from the complementary advantages of both individual models}. This \rebuttal{merging} layer 
combines the outputs from \dsim{} and \qsim{} using a merging parameter \rebuttal{$\alpha$, to control the contribution of the individual models}. 
We investigated 
\rebuttal{empirical} score-based and rank-based merging strategies. 
The score-based model \rebuttal{assumes a linear combination of} 
both respective \rebuttal{\emph{scores}} 
$r_i^{\text{\dsim}}$ and $r_i^{\text{\qsim}}$ from \dsim{} and \qsim{}, \rebuttal{in which their individual contribution is controlled by the hyperparameter $\alpha$:}
%
        %
         \[
        r_i^{\text{\dqsimscore}}(d) = \alpha\,r_i^{\text{\dsim}}(d) + (1-\alpha)\,r_i^{\text{\qsim}}(d) 
        \]
%
\thms{The rank-based model combines the distractor \rebuttal{\emph{ranks}} $\rho_{i}^{\text{\dsim}}$ and $\rho_{i}^{\text{\qsim}}\in\{1, 2, 3, ..., N\}$ from \dsim{} and \qsim{} into the score}
 \[ 
 r_i^{\text{\dqsimrank}}(d) = \frac{\alpha}{\log \left ( \rho_{i}^{\text{\dsim}}(d) + 1 \right )} + \frac{1-\alpha}{\log \left ( \rho_{i}^{\text{\qsim}}(d) + 1 \right )}
\]        
\newtext{This scoring function is based on weighted combination of inverse distractor rankings, \rebuttal{such that} high-ranked distractors have more weight. We use \rebuttal{logarithmic smoothing} 
to avoid the potential contribution of low-ranked distractors from \rebuttal{vanishing too rapidly}. 
} 

\par 

\subsection{Training}
\label{sec:training}

We use \emph{contrastive learning} as our training \rebuttal{strategy~\cite{sun2022dual}}. \rebuttal{Contrastive learning~\cite{mikolov2013efficient,chopra2005learning,smith2005contrastive}} is a machine learning technique that aims to learn representations of data by contrasting similar and dissimilar examples. It aims to bring similar instances closer together in the representation space by maximizing the similarity between their embeddings, while pushing dissimilar samples further apart by minimizing their similarity.
 
In a contrastive learning setting, it is often the case that similar \thms{example pairs} (\ie also called positive examples) are 
available explicitly in training datasets, whereas dissimilar or negative examples need to be sampled from an extremely large pool of instances. 
\thms{For the \qsim{} model, a positive pair consists of two questions sharing at least one distractor, whereas for the \dsim{} model, we require similar representations for a given (stem, key) item and a distractor corresponding to the same \mcq.}
 
As a negative sampling strategy, we use in-batch negatives~\cite{karpukhin2020dense} while training our models. For \dsim, the in-batch negatives are gold-standard positive distractors for the other instances in the same batch. While for \qsim{}, the in-batch negatives are the positive questions that come from the other instances in the same batch. Reusing gold standard distractors or questions from the same batch as negatives \thms{makes training more efficient, compared to randomly sampling negatives for each positive pair in the batch.}
 
\thms{With the notation $r_i(d)$ (common in both \dsim{} and \qsim{}) for scoring \mcq$_i$ against distractor $d$, and by introducing the sigmoid function $\sigma(r)=1/(1+e^{-r})$, we can write the contrastive loss\rebuttal{\cite{sohn2016improved}} $\mathcal{L}_i$ to be minimized for \mcq$_i$ with matching distractors $d^+$ as follows:}
\thms{
\[
\mathcal{L}_i = -\sum_{d^+} \log \sigma\big(r_i(d^+)\big) - \sum_{d^-} \log \sigma\big(-r_i(d^-)\big)
\]
in which $r_i(d^+)$ denotes the score of a positive distractor for the considered question, and $r_i(d^-)$ the scores for the in-batch negatives (summed over the considered batch of training instances). 
If the quantity $\sigma\big(r_i(d)\big)$ is interpreted as the probability that distractor $d$ is compatible with \mcq$_i$ (in the sense of model \dsim{} or \qsim{}), then minimizing the above loss term can be understood as maximizing the joint estimated probability of $d^+$ being compatible distractors for \mcq$_i$, and the in-batch negatives $d^-$ to be incompatible ones.  
}

\subsection{Using the models for predictions}

\newtext{This section describes the inference mechanism for our models. Inference refers to using a trained model 
to make predictions about new data. 
For each of the models, the goal is inducing an ordering of all candidate distractors in response to a given question stem and answer key, such that the top ranked ones can be proposed as fitting distractors.}

For the \dsim{} model, since the \thms{considered (stem, key) pair and the distractor to be scored against it} 
are independently fed to the network, the embeddings of the pool of distractors can be computed offline. \thms{The vector representation $h^{(sk)}$ of a given stem and its answer key is calculated, compared through the dot product with each of the pre-calculated distractor representations $h^{(d)}$, and these are then ordered according to decreasing score.}
 
Similarly, for the \qsim{} model, the pool of questions' embeddings is \thms{calculated offline and stored. 
At run time, for a given question stem $s$, we compute its embedding $h^{(s)}$, score it against all pre-calculated stem representations for the \mcq{}s in the corpus, and rank the candidate distractors according to the decreasing score of their corresponding question stem. Note that we assign that same score to each of the distractors of a given stem (for use in \dqsimscore). We then rank all distractors according to decreasing scores (randomly ordering those with identical scores). } 
        
Finally, \thms{once the scores for \dsim{} and \qsim{} are calculated for each candidate distractor, the \dqsim{} model can be evaluated directly, by ranking them according to the decreasing score $r^{\text{\dqsimscore{}}}$ or $r^{\text{\dqsimrank{}}}$.}



\section{Experimental Design}
\label{sec:experimentaldesign}

\newtext{This section describes the evaluation methodology 
and the metrics we used to measure the quality of the generated distractors using the different methods described in \secref{sec:methodology}. \secref{sec:evaluation_setup} introduces our hypotheses and the experiments we designed to test them. 
The automatic evaluation metrics we used are explained in \secref{sec:automated_metrics}.}

\subsection{Evaluation Setup}
\label{sec:evaluation_setup}
\newtext{In order to validate our models' theoretical effectiveness and practical applicability, we formulate the following three key hypotheses}
\chris{, which we will test through experiments based on both automatic and human annotator evaluation:}
\begin{enumerate}[label=\textbullet~\textbf{Hypothesis~\arabic*}:, align=parleft, labelindent=\parindent,leftmargin=*,labelwidth=\widthof{Hypothesis~0},
itemindent=\widthof{Hypothesis~0},ref=Hypothesis~\arabic*]
    \item \label{it:hypo1} \emph{Context-aware models, based on generic pre-trained language models, lead to more effective
distractor selection models than shallow prediction models based on manually engineered features.} 
    \item \label{it:hypo2} \emph{Manual distractor quality scores are correlated with machine-generated distractor candidate rankings.}
    \item \label{it:hypo3} \emph{Top-ranked machine-proposed distractor candidates are comparable in quality to expert-generated distractors, for a given question stem and answer key.}
\end{enumerate}

    

\chris{For \ref{it:hypo1}, we first of all set up a}
large-scale automatic evaluation experiment with the Televic dataset (see \tabref{tab:dataset})
\chris{.}
In addition, a focused small-scale automatic evaluation of context-aware and feature-based models was carried out on the WeZooz external data (see \secref{sec:wezoozdataset} for details) that contains several subjects. 

\chris{We complemented that automatic evaluation with human evaluation, since hard comparison} 
of ground-truth distractors with machine-generated distractors may not give the whole picture of accuracy.
\chris{Indeed, both for} 
language learning and factual knowledge learning, {\mcq}s can have a potentially large set of viable distractors that are not included by the gold standard distractor set.
\chris{Thus,} automated metrics could flag a correctly proposed candidate distractor as wrong because of the scarcity of the gold standard dataset.
To avert this problem, many previous works asked human experts to judge the quality of the distractors that were generated by their systems~\cite{singh2013automatic, araki2016generating}.
\chris{Hence, we} 
also invite\chris{d} teachers to provide their expert opinion, \thms{each focusing solely on a set of questions on their own subject of expertise.} 
\thms{In the following paragraphs}, we explain the procedure we followed to set up 
\chris{that} expert evaluation
\chris{, which we will use in assessing all aforementioned Hypotheses 1--3.}

First, we prepared a small sample of test questions for language and factual knowledge learning.
For language learning, we used French and English.
These questions were randomly drawn from the held\chris{-}out test split of the Televic dataset introduced in \secref{sec:televicdataset}.
For the factoid type questions, we use the WeZooz dataset introduced in \secref{sec:wezoozdataset}.
Each of the subjects contains a fixed list of 50 questions.
Second, we applied the different trained models to rank distractors according to their relevance for each question in the test set.
We then kept the top-10 ranked candidate distractors for each of the models.
Finally, teachers were 
shown distractor predictions 
\chris{unified over} all models (\ie duplicates were removed) 
\chris{as well as} the provided gold-truth distractors for each test question \chris{(see the illustration provided}
in \figref{fig:screenshot} in \appref{appendix:annotationplatform}\chris{)}.
\chris{Note that the order of the unified list of distractors was randomized, to avoid introducing order bias.}


The teacher participants were explicitly instructed to rate each candidate distractor based on how much they thought it would help them if they were given the task of preparing distractors for that specific question. 
Specifically, we asked them to annotate each distractor independently of the other distractors in the list\chris{,} based on a four-level annotation scheme that we designed to measure the quality of distractors.
\chris{Our scale} is closely related to the three-point \thms{evaluation scale} 
proposed by \cite{araki2016generating} 
(\Tabref{tab:annotationschemeexamples} shows examples of each category):

\begin{itemize}
    \item \emph{True Answer}: specifies that the distractor partially or completely overlaps with the \thms{answer} key.
    \item \emph{Good distractor}: specifies that the distractor is viable and could be used in an MCQ as is. 
    \item \emph{Poor distractor}: specifies that the distractor is on topic but could easily be ruled out by students. This could happen due to one or both of the following reasons. 
      
    \begin{itemize}
   
        \item \emph{Poor meaning}: the distractor has poor meaning. For example, it is too general, \thms{although not completely off-topic}.  
        \item \emph{Poor format}: the distractor's format is different from the format of the answer \thms{key} and \thms{does} 
        not fit with the stem.  
    \end{itemize}

    \item \emph{Nonsense distractor}: specifies that the proposed distractor is completely out of context. 
\end{itemize}

Although the third category (\ie poor distractor) implies that the proposed distractor is ineffective as is, a minor tweak \thms{may} result in a useful distractor. Furthermore, even if a significant change is required, it may inspire teachers 
to create new effective distractors.

Using the annotations we gathered from the teachers, we tested 
\chris{Hypotheses} 2 and 3.
For \emph{\ref{it:hypo2}}, 
\chris{we evaluated whether the higher ranked 
distractors also have a higher perceived usefulness.}
This was done 
by comparing the
\chris{human scoring of distractor candidates in the top-5 to that of those ranked 5--10:}
\chris{for a good distractor generation model, the top-5 should on average contain significantly more `good' ones.}
We designed a statistical analysis to test the null hypothesis that the rating distribution is not related to whether candidate distractors were ranked top-5 or 5--10.
We used Fisher's exact test\footnote{\semere{We also conducted a chi-square test and reached the same conclusions.}} to test this hypothesis.

For \emph{\ref{it:hypo3}}, we evaluated the extent to which the teachers perceived the system-generated distractor candidates as the ground-truth distractors. Again, we use Fisher's exact test to test the null hypothesis that the distribution of quality of distractors is not related to whether the distractors are human-generated or system-generated.

\subsection{Automated Metrics}
\label{sec:automated_metrics}

 We use two groups of information retrieval metrics \lucas{to automatically evaluate} our systems:
\begin{enumerate*}[(1)]
  \item Order-unaware metrics: Recall@$k$ and Precision@$k$\chris{,} which measure the fraction of gold-standard distractors that are in the top-$k$ distractors and the fraction of relevant distractors in the top-$k$ retrieved distractors, respectively.
  \item Order aware metrics: mean reciprocal rank (MRR) and mean average precision (MAP)\chris{,} which \chris{respectively reflect} 
  \chris{how high the most relevant item is ranked in the list, and how high all relevant ones are ranked on average.}
\end{enumerate*}

\begin{table*}[ht]
\centering
\footnotesize
\caption{Annotation scheme examples 
}
\label{tab:annotationschemeexamples}

\begin{tabular}{p{3cm} p{1.5cm}p{2.0cm}p{2.0cm}p{3cm}}
\toprule
\textbf{Question} & \textbf{Answer}&   \textbf{Distractors}   &    \textbf{Category}     & \textbf{Moderation}                                                                              \\ \midrule
\multirow{4}{3cm}{Which inhabitants are not happy with Ethiopia's plans of the Nile?}    & \multirow{4}{2em}{Egyptians}  &  1. Itali  &  Poor format & because of wrong spelling.                                                                  \\
                        &   & 2. Kenyans  & Good  & - \\
                        &  & 3. gypsies    & Poor meaning & because 
                           
                           \\ \midrule
\multirow{3}{3cm}{My mum brought the washing in .... it was raining.}    & \multirow{3}{2em}{because}  &  1. until  & Good &  -                                                                  \\
                        &   & 2. since    & True Answer     & \\
                        &  & 3. investigate & Nonsense  & out of context
                           
                           \\ \midrule

\multirow{4}{3cm}{How old was Beethoven when he died?}    & \multirow{4}{5em}{56 years}  &  1. 1.5v & Nonsense &  out of context                                                                  \\
                        &   & 2. 60 years  & Good & -  \\
                        & & 3. 180 years  & Poor meaning & humans cannot live 180 years.                                     \\ \bottomrule                                     
\end{tabular}
\end{table*}

\section{Results and Discussion}
\label{sec:resultsanddiscussion}

In this section, we provide evidence of the effectiveness of our context-aware models by reporting the experimental results and discussing the insights gained. \secref{sec:automaticevaluation} compares the baseline with our proposed context-aware models using reproducible automated metrics \chris{(\ref{it:hypo1})}.
\secref{sec:human_evaluation} discusses the user study results with experts \chris{(Hypotheses~1--3)}.
Note that all the \chris{numerical} results reported in this section are in percentage 
\chris{points.}

\subsection{Automatic Evaluation}
\label{sec:automaticevaluation}

When considering the results of our automated evaluation based on the recovery of ground-truth distractors, it is essential to note that information about ground-truth distractors for a given item was never used during the model's training. 
\Tabref{tab:results} shows the large-scale evaluation of the systems on the Televic \thms{test set}. \rebuttal{We report our results as the mean and standard deviation of five different runs of our models using five random seeds as shown in \tabref{tab:results}.}
All three context-aware models consistently outperform our \thms{feature based model (denoted `baseline')} 
on all metrics.
\dqsim{} performs the best \thms{according to most} metrics, 
\chris{confirming} that \qsim{} and \dsim{} have their own \thms{(complementary)} merits.
{\qsim} is better than {\dsim} at recovering ground truth distractors (\ie Recall@10 of \rebuttal{82.3} compared to \rebuttal{76.0}), but inferior at ranking the best relevant distractor \rebuttal{at the top} in the list\chris{, which we conclude from the lower Precision@1}
\rebuttal{(40.4 vs.~44.9) and MRR (55.6 vs.~60.7) scores.
This is related to the nature of the {\qsim} model. The candidate distractors belonging to its best matching question would be put at the top of the returned distractors in a random order. 
Our results show that {\dsim} is better at estimating the most likely distractor than {\qsim} is in finding a relevant question \emph{and} arriving with the relevant distractor on top after random ordering. However, the Precision@4 results show that {\qsim} has more success in identifying a question with good distractors, than {\dsim} has in detecting good distractors among its top 4 results. The other reported metrics (Recall@10, Precision@4, MAP) indicate the overall higher effectiveness of {\qsim} when looking further than only the top result.
}
\chris{In our MCQ generation setting, recall \thms{within the top 10 results} is the more important metric, since the presence of high quality distractors in the automatically generated list is more important than their correct ranking.}

\begin{table}[H]
\footnotesize
\begin{center}
\caption{Automatic ranking evaluation Full-ranking}
\label{tab:results}
\begin{tabular}{lllllll}
\toprule
\textbf{Models} & 
\multicolumn{1}{l}{\textbf{R@10}} &
\multicolumn{1}{l}{\textbf{P@1}} &
\multicolumn{1}{l}{\textbf{P@4}} &
\multicolumn{1}{l}{\textbf{MAP}} &
\multicolumn{1}{l}{\textbf{MRR}} &
\\\midrule
Baseline    & \rebuttal{71.3\std{1.2}}  & \rebuttal{21.1\std{1.8}}              & \rebuttal{23.7\std{0.5}} & \rebuttal{33.5\std{1.0}}    & \rebuttal{43.9\std{1.9}}\\
\dsim       & \rebuttal{76.0\std{0.7}}  & \rebuttal{\textbf{44.9\std{0.5}}}     & \rebuttal{24.4\std{0.8}} & \rebuttal{44.9\std{0.6}}    & \rebuttal{60.7\std{1.3}} \\
\qsim       & \rebuttal{82.3\std{0.5}}  & \rebuttal{40.4\std{1.5}}              & \rebuttal{35.9\std{0.9}} & \rebuttal{54.9\std{0.9}}    & \rebuttal{55.6\std{1.1}}\\
\dqsim      & \rebuttal{\textbf{91.7\std{0.6}}} & \rebuttal{41.9\std{0.8}} & \rebuttal{\textbf{38.2\std{0.7}}} & \rebuttal{\textbf{57.3\std{0.5}}} & \rebuttal{\textbf{62.8\std{0.4}}}\\
\bottomrule
\multicolumn{1}{c}{} & \\
\multicolumn{6}{l}{\footnotesize R:recall, P: precision, MAP: mean avg.~ precision,} \\
\multicolumn{6}{l}{\footnotesize MRR: mean reciprocal rank; evaluation on Televic test set.}
\end{tabular}
\end{center}

\end{table}

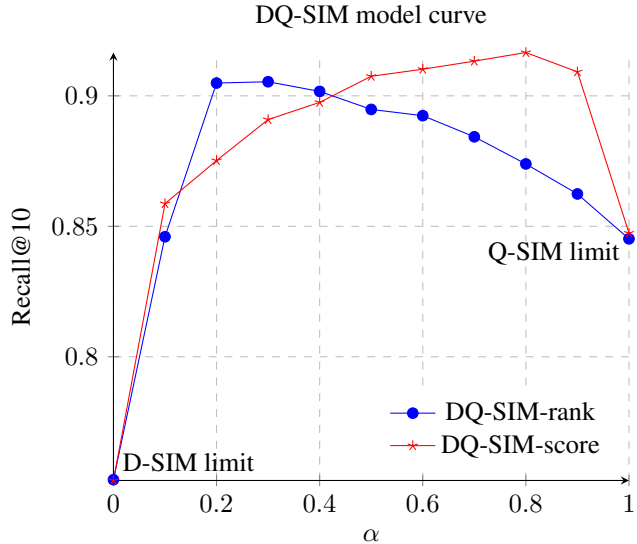
\begin{figure}[ht]
\centering
\begin{tikzpicture}
\begin{axis}[
legend style={draw=none},
    axis lines = left,
    title={DQ-SIM model curve},
    xlabel={ $\alpha$},
    ylabel={Recall@10},
    xmin=0, xmax=1,
    ytick={0.70,0.75,0.80,0.85,0.90,0.95,1.0},
    legend pos=south east,
    ymajorgrids=true,
    xmajorgrids=true,
    grid style=dashed,
]

\addplot[
    color=blue,
    mark=otimes*,
    ]
    coordinates {
    (0.0, 0.7529)
(0.1, 0.846)(0.2, 0.9049)(0.3, 0.9054)(0.4, 0.9017)(0.5, 0.8948)(0.6, 0.8924)(0.7, 0.8843)(0.8, 0.8739)(0.9, 0.8624)(1.0, 0.8452)
    };

\addplot[
    color=red,
    mark=star,
    ]
    coordinates {
(0.0, 0.7526)(0.1, 0.8587)(0.2, 0.8752)(0.3, 0.8909)(0.4, 0.8975)(0.5, 0.9075)(0.6, 0.9102)(0.7,0.9133)(0.8, 0.9166)(0.9, 0.9092)(1.0, 0.8474)
    };
\node at (axis cs:0.0, 0.7526) [anchor=south west] {D-SIM limit};
\node at (axis cs:1.0, 0.8474) [anchor=north east] {Q-SIM limit};
 \legend{DQ-SIM-rank, DQ-SIM-score}
\end{axis}
\end{tikzpicture}
\caption{Different $\alpha$ values for combining \qsim{} and \dsim{} models using rank and raw scores on the validation set}
\label{fig:alpha}
\end{figure}

\Figref{fig:alpha} depicts the performance of {\dqsim} \chris{for the two merging strategies,} in terms of Recall@10 
on the validation set described in \secref{subsection:dqsim}.
The linear combination of the scores outperforms the rank-based merging strategy.
The score-based strategy achieves the best performance at $\alpha =$ 0.8, giving more weight to the {\qsim} model. \newtext{This is reasonable given that the {\qsim} model outperforms the \dsim{} model on the recall metric.}




\Tabref{tab:results_subjects} compares the baseline with \dqsim{} (\ie \chris{the} best context-aware model according to \chris{the evaluation on the} Televic dataset
) in a small-scale setting for all 
\thms{four Wezooz dataset subjects as well as English and French from the Televic test set}.
Since 
\chris{we want} to compare 
\chris{models in terms of} their ability to rank relevant distractors higher in the list, we added the ground-truth distractors from all the subjects to the existing distractors pool.
Ideally, the best model would rank all the ground\chris{-truth} distractors high in the list.
Similar to the large-scale evaluation, {\dqsim} consistently outperforms the baseline for all subjects on both metrics.
Recall@10 and MAP are higher for the language category than 
\chris{for factoid questions} because the test questions for the former come from the same distribution (\ie Televic test questions) as \thms{the data the models were trained on (\ie Televic train set)}. 
On the other hand, the test data for the factoids come from a different distribution (\ie WeZooz dataset) than the training data
such that the evaluation for these subjects additionally measures the robustness of the model to a data distribution shift (\ie its domain transfer abilities). \newtext{The \dqsim{} model is far more robust than the baseline.}


\begin{table}[h!]
\begin{center}
\caption{Small scale Automatic ranking evaluation }
\label{tab:results_subjects}
\begin{tabular}{lccccc}
\toprule
\textbf{Models} & 
\multicolumn{2}{c}{\textbf{Baseline}} &
\multicolumn{2}{c}{\textbf{\dqsim}} &
\\\cmidrule(lr){2-3} \cmidrule(lr){4-5}

& \multicolumn{1}{c}{\textbf{R@10}} &
\multicolumn{1}{c}{\textbf{MAP}} &
\multicolumn{1}{c}{\textbf{R@10}} &
\multicolumn{1}{c}{\textbf{MAP}} 

\\\midrule
English\textsuperscript{*}     & 60.1  & 33.6          & \textbf{98.3}     & \textbf{85.8} \\ 
French\textsuperscript{*}      & 46.6  & 17.7          & \textbf{81.1}     & \textbf{61.1} \\  \midrule
Nat. Sciences  & 24.3  & 7.7           & \textbf{74.3}      & \textbf{37.3} \\ 
History     & 14.3  & 3.4           & \textbf{62.2}      & \textbf{35.7}   \\ 
Biology     & 30.6  & 7.6           & \textbf{72.0}      & \textbf{41.8}  \\   
Geography   & 32.3  & 12.1          & \textbf{61.5}      & \textbf{34.4}  \\
\bottomrule
\multicolumn{1}{c}{} \\
\multicolumn{5}{l}{\footnotesize{R: recall, MAP: mean avg.~precision; * denotes subject }}  \\
\multicolumn{5}{l}{\footnotesize{is drawn from the Televic test set, while the rest are from WeZooz.}} 
\end{tabular}
\end{center}

\end{table}




\subsection{Expert Evaluation}
\label{sec:human_evaluation}

Following the procedure introduced in \secref{sec:experimentaldesign}, a total of 12,723 ratings for distractor quality were gathered from the annotations by teachers (see \tabref{tab:annotationsdatadescription} for details of rating statistics).
These ratings come from the top\chris{-}10 ranked distractors for each of the four models, and the ground-truth distractors (\ie all simultaneously presented and randomly shuffled). \thms{We retained the gold standard distractors in the lists to be annotated,}
because we wanted to investigate the agreement among teachers in creating distractors.
In the following subsections, we 
study teachers' (dis)agreement on the quality of distractors, compare the various models using the evaluation from experts, and \thms{revisit Hypotheses~1--3 in light of these results.} 


\vspace{\baselineskip}
\subsubsection{Inter-annotator agreement}
\label{sec:interannotatoragreement}
\chris{We adopt 2 strategies to assess inter-annotator agreement.}
First, we 
\chris{analyze} how teachers rated the ground-truth distractors, which were made by other teachers who prepared the questions.
As can be seen from \tabref{tab:groundtruthconsistency}, in general, we find that 79\% of the actual distractors were deemed good, 11\% poor, followed by 7\% nonsense and 3\% true answers.
There is greater agreement between teachers in what is considered a good distractor on the factoids than for language \thms{learning exercises}  
(83\% \vs 70\%).     

\begin{table}[t]

\begin{center}
\caption{Inter-annotation agreement of ground-truth distractors (\%)}
\label{tab:groundtruthconsistency}
\begin{tabular}{lcccc}
\toprule
 & 
\multicolumn{1}{c}{\textbf{True Ans.}} &
\multicolumn{1}{c}{\textbf{Good }} &
\multicolumn{1}{c}{\textbf{Poor}} &
\multicolumn{1}{c}{\textbf{Nonsense }} 

\\\midrule
Languages & 5 & 70 & 14 & 11 \\
Factoids &  2 & 83 & 9 & 6  \\
\midrule
Overall &  3 & 79 & 11 & 7  \\
\bottomrule
\end{tabular}
\end{center}

\end{table}

Second, we study the agreement of teachers by asking them to rate the same set of distractors using 
\chris{our} four-level scale annotation scheme
. 
\chris{We selected the subjects}
English, from the languages category, and History, from factoids\chris{,} 
for annotations by at least two teachers. \Tabref{tab:interannotation_jaccard} shows the inter-annotator agreement of teachers using the Jaccard similarity coefficient.
The Jaccard similarity measures similarity between two sets of data by calculating 
\chris{what fraction of the union of those datasets is covered by their intersection.}
In our case, it is calculated as 
the number of times the teachers agreed on a distractor category label (\ie one of the four quality labels) divided by the total number of distractors that were annotated \chris{(by either annotator)} with that label. 
In general, 
\chris{we note} a higher agreement on what is considered a good distractor and a nonsense distractor.
Particularly, the overall agreement between the History teachers is higher than the English teachers. 
This is in line with the higher agreement 
\chris{for} factoid type questions discussed in the previous paragraph. 
The Jaccard similarity is sensitive to small sample sizes.
For example, a total of only two distractors were rated `true answer' by the history teachers which yielded no similarity (\ie a `0' in the first column in \tabref{tab:interannotation_jaccard}).

\chris{Calculating the inter-annotator agreement with the commonly used Cohen's kappa~\cite{mchugh2012interrater} value, we confirm aforementioned higher agreement for factoid questions than for language: Cohen's kappa is 29.3 among English teachers, which represents ``fair agreement'', and 40.5 among History teachers, indicating ``moderate agreement''.}

\chris{As a final metric to assess potential ambiguity in scoring distractors,}
\chris{we calculate conditional probabilities $P(X|Y)$ of having a second annotator assigning label $X$ given that a first one said $Y$.}
For example, unsurprisingly, the probability of rating a distractor `good' given that it was rated `nonsense' by 
\chris{another} teacher and vice-versa was 6\% for English and 5\% for History.
This implies that the confusion in differentiating good distractors from nonsense distractors was minimal. Details are presented in \tabref{tab:conditionalprobraters} in \appref{appendix:userstudydetails}.




\begin{table}[t]

\begin{center}
\caption{Inter-annotation agreement of experts in terms of Jaccard similarity coefficent (\%)}
\label{tab:interannotation_jaccard}
\begin{tabular}{lccccc}
\toprule
\textbf{Subjects} & 
\multicolumn{1}{c}{\textbf{True}} &
\multicolumn{1}{c}{\textbf{Good}} &
\multicolumn{1}{c}{\textbf{Poor}} &
\multicolumn{1}{c}{\textbf{Nonsense}} & 
\multicolumn{1}{c}{\textbf{Overall}} 
\\\midrule
English & 25.8 & 42.9 & 12.8 & 40.0 & 47.9\\
History & 0.0 & 43.6 & 24.3 & 59.7  & 57.7\\

\bottomrule
\end{tabular}
\end{center}

\end{table}


\vspace{\baselineskip}
\subsubsection{Evaluation of models by experts}
\label{sec:model_comparison}

\Tabref{tab:results_expert_eval} shows the expert evaluation of distractors in terms of \emph{good distractor rate} (GDR@10) and \emph{nonsense distractor rate} (NDR@10). 
GDR@10 is calculated as the percentage of distractors that were rated `good' among the top 10 ranked distractors for each model. Similarly, NDR@10 is calculated as the percentage of distractors that were rated `nonsense' among the top 10 ranked distractors for each model.
We are interested in reporting the NDR metric because \begin{enumerate*}[(i)]
    \item it could be used to distinguish between good and bad systems, and
    \item in a real-world scenario discarding a system with high NDR score could be helpful since \thms{the frequent occurrence of nonsense distractors may} scare away users by eroding their trust in \thms{the model}.
\end{enumerate*}
The reported metrics 
are averages of all the subjects in each category. $\uparrow$ indicates larger values are better and $\downarrow$ indicates smaller values are better.
In general, context-aware models were rated better in proposing plausible distractors than the baseline model.
They also produced 
\chris{fewer} nonsense distractors. 
The {\dqsim} outperformed all the other models.
On average, 3 out of its top 10 proposed distractors were rated good distractors. Moreover, \thms{on average 5.5}  distractors for languages and 5 for factoids were generally found on-topic (\ie distractors rated as either good or poor distractors) for \dqsim.

The NDR@10 is lower for all models 
\chris{for language subjects} than 
\chris{for factoid questions}.
We hypothesize this is 
because the test data for the language category comes from the same distribution the models were trained on.


\begin{table}[t]
\begin{center}
\caption{Expert evaluation of distractors (\%)}
\label{tab:results_expert_eval}
\begin{tabular}{lcccc}
\toprule
\textbf{Models} & 
\multicolumn{2}{c}{\textbf{Language learning}} &
\multicolumn{2}{c}{\textbf{Factoid learning }} 
\\\cmidrule(lr){2-3} \cmidrule(lr){4-5}
& \multicolumn{1}{c}{\textbf{GDR@10 $\uparrow$}} &
\multicolumn{1}{c}{\textbf{NDR@10 $\downarrow$}} &
\multicolumn{1}{c}{\textbf{GDR@10 $\uparrow$}} &
\multicolumn{1}{c}{\textbf{NDR@10 $\downarrow$}} 

\\\midrule
Baseline & 23.6 & 45.4 & 13.6 & 66.0 \\ 
\dsim & 25.9 & 45.2 & 15.0 & 64.8 \\ 
\qsim & 26.3 & 45.3 &  19.0 & 61.6 \\ 
\dqsim & \textbf{27.9} & \textbf{44.6} & \textbf{28.9} & \textbf{50.1}   \\ 

\bottomrule
\multicolumn{1}{l}{} \\
\multicolumn{5}{l}{\footnotesize GDR: good distractor rate, NDR: nonsense distractor rate;} \\
\multicolumn{5}{l}{\footnotesize $\uparrow$: higher is better, $\downarrow$: lower is better; evaluation on WeZooz test set } \\
\end{tabular}
\end{center}

\end{table}

\vspace{\baselineskip}
\subsubsection{Discussion 
\chris{of} key hypotheses}
\label{sec:hypotheses_testing}
\chris{We now discuss to what extent our experimental results confirm our aforementioned key Hypotheses 1--3.}

\ref{it:hypo1} states that the context-aware models generate better quality distractors than the feature-based models. As discussed in \secref{sec:automaticevaluation}, the automated evaluation shows that the context-aware models consistently outperform the feature-based model on the Televic and WeZooz datasets. The human evaluation in \secref{sec:model_comparison} 
\chris{further confirms this} by demonstrating that distractors generated by context-aware models were rated higher in quality than those generated by feature-based models. 

\newtext{
\ref{it:hypo2} states that human distractor quality ratings are correlated with the automated candidate distractor rankings.}
To test this hypothesis, we collapsed the four ratings into two categories: \emph{plausible} (\ie rated as good distractors) and \emph{less plausible} (\ie rated as true answer, bad and nonsense distractors). 
\Tabref{tab:contingency_table_hypo2} in \appref{appendix:userstudydetails} shows the 
contingency table \chris{for Fisher's exact test} for 
\chris{our} best model\chris{, \ie} {\dqsim}.
\chris{The fraction of} top\chris{-}5 ranked distractors that received `good distractor' ratings (\ie 30.3\%) is higher than 
\chris{that for the ones ranked 5--10} (\ie 21.6\%). 
 We found that this difference 
 is statistically significant. Indeed, the null hypothesis that the automatic ranking of distractors is unrelated to how teachers rated them is \thms{strongly rejected ($p=$1.7e-8)}. 

\newtext{
\ref{it:hypo3} asserts that the quality of top-ranked machine-generated distractors is comparable with human-made distractors.}
To test 
\chris{this,} we compare the distribution of ratings of the ground-truth distractors (\ie expert-generated distractors) with the distribution of ratings 
\chris{for the} \dqsim{} model (\ie system-generated distractors).
As 
\chris{for \ref{it:hypo2},} we collapse the ratings into \emph{plausible} and \emph{less plausible} classes.
\Tabref{tab:contingency_table_hypo3} in \appref{appendix:userstudydetails} shows the 
contingency table \chris{for Fisher's exact test, to compare the quality} 
between system-generated and human-generated distractors.
The null hypothesis that the source of the distractor (\ie human-generated or system-generated) is unrelated to the quality label assigned by the teachers\chris{,} is \thms{strongly rejected ($p<$1.e-10)}. 
\thms{Indeed, } 
the quality of the human-generated distractors was found to be better than the system-\thms{proposed} distractors.
\chris{Still, we believe system-generated distractors have value: given that they can be generated quickly and automatically, presenting them as suggestions --- rather than relying on a fully automated system --- seems a practically meaningful way of working, which could save teachers a significant amount of time (compared to purely creating a list of distractors without any assistance).}


\section{Conclusion and Future Work}
\label{sec:conclusionandfuturework}

This paper introduced and evaluated multilingual context-aware distractor retrieval models for reusing distractor candidates that can facilitate the task of MCQ creation. Particularly, we proposed three models: \begin{enumerate*}[(1)]
    \item \chris{The} {\dsim} model that learns similar contextual representations for similar distractors, 
    \item \chris{The} \qsim{} model that requires similar questions to have similar representations, and
    \item \chris{The} \dqsim{} model that linearly combines the previous two models benefiting from their respective strengths. \thms{Importantly, the \dqsim{} model showed a considerably reduced nonsense distractor rate, which we consider a useful asset in terms of trust in the model by teachers.}
\end{enumerate*}
\rebuttal{We also asked teachers to evaluate the quality of distractors using a four-level annotation scheme that we introduced.}
\rebuttal{As the result, }teachers \thms{considered} 3 out of 10 suggested distractors as high-quality, to be readily used. 
\rebuttal{Additionally,} they found two more distractors to be within topic, albeit of 
\chris{lower} quality, and \thms{useful as inspiration for} 
teachers to \thms{come up with their own good} distractors. \rebuttal{Finally, we released a test consisting of 298 educational \mcq{}s with annotated distractors covering six subjects and a 77K distractor vocabulary to promote further research.}

In future work, we foresee three directions.
First, it is worth reiterating that the current work assumes access to a substantial pool of distractors.
Even though with such large item pools, it is expected that many options are available for an incoming newly written question, the current work is unable to generate a brand new distractor. 
A possible solution 
could be to employ pure generative models that can freely generate distractors.
Moreover, generative models could correct the `poor format' errors.
However, it has to be noted that such models require access to a context where the distractors and questions come from, such as a chapter of a book, Wikipedia article, etc. 
A~second research direction is to extend the current work to a multimodal system that considers other sources of information\chris{, \eg} 
images that accompany MCQs in digital learning tools.
Finally, an area that we are currently investigating is how to 
\chris{make sure the complete list of distractors in a single MCQ is sufficiently diverse:}
\chris{note that in} 
the present study, we were only interested in retrieving a list of plausible distractors independent of each other.
However, typical MCQ distractors 
\chris{should} not only \chris{be} plausible but also \thms{sufficiently} diverse.

\clearpage

\begin{ack}
\newtext{This work 
\chris{was} funded by VLAIO (`Flanders Innovation \& Entrepreneurship') in Flanders, Belgium, through the \emph{imec-icon} project AIDA (`AI-Driven e-Assessment'). This research also received funding from the Flemish Government under the “Onderzoeksprogramma Artificiële Intelligentie
(AI) Vlaanderen” programme.
We would like to thank the AIDA partners  Televic Education and  WeZooz Academy for contributing data and use cases, as well as the ZAVO (`Zorgzaam Authentiek Vooruitstrevend Onderwijs') secondary school teachers for participating in the study.}

\end{ack}

\bibliography{references}

\appendix

\section{Training and Implementation details}
\begin{itemize}
    \item \textbf{Feature-based models}: for feature extraction and model training we use components from the scikit-learn package for python \cite{pedregosa2011scikit}. As negative training examples, we sample a total of 100 non-distractors for each MCQ.
    \item \textbf{Context-aware models}: our transformer based model is implemented using Pytorch \cite{NEURIPS2019_9015} and Huggingface \cite{wolf2019huggingface}. We initialize our encoder with \texttt{bert-base-multilingual-uncased}. We fine-tune the last two layers and leave the other layers frozen. \rebuttal{The most important hyper-parameters are the learning rate, batch size, the duration of training, and the output width of our dense layer. To avoid extensive hyper-parameter tuning, we made the following choices. First,} we choose the output dimension of the dense layer to be $d_{out}=700$ \rebuttal{because we empirically found that it yielded good results. For the learning rate, we kept the choice of $10^{-5}$ from Karpukhin~\etal \cite{karpukhin2020dense} in combination with the robust Adam optimizer~\cite{kingma2014adam}. Also in line with \cite{karpukhin2020dense}, we know increasing batch size may lead to slightly improved results, and thus decided the batch size to be 64, the highest value that would fit our V100 memory. We train each model for
25 epochs at which point performance on the development begins to plateau due to overfitting.}
\end{itemize}

\section{Feature Vector Description}
\label{appendix:featureset}

We describe each feature we used to build our feature-based classifiers in below.

\begin{enumerate}
\footnotesize  
\item \texttt{tfidf\_word\_match\_share} : a word overlap metric between both \emph{k} \& \emph{d} and \emph{s} \& \emph{d} which weighs overlapping words according to their inverse document frequency value.  
\item \texttt{word\_match\_share}: fraction of word tokens that are shared between both \emph{k} \& \emph{d} and \emph{s} \& \emph{d}. 
\item \texttt{equal\_num} : boolean feature that checks whether \emph{k} \& \emph{d} have equal numbers of digits. 
\item \texttt{longest\_substring} : fraction of longest matching sub-string between \emph{k} and \emph{d}. 
\item  \texttt{token\_len\_sim} : boolean feature that checks if the amount of tokens in \emph{k} is equal with \emph{d}.          
 \item  \texttt{token\_len\_diff} : difference in amount of tokens in \emph{k} and \emph{d}.         
\item    \texttt{char\_len\_sim} : boolean feature that checks if the amount of characters in \emph{k} is equal with \emph{d}.
 \item  \texttt{char\_len\_diff} : difference in amount of tokens in \emph{k} and \emph{d}.         
\item   \texttt{is\_caps} : boolean feature that checks if both \emph{k} and \emph{d} are capitalized.
 \item  \texttt{count\_caps} : boolean feature that checks if both \emph{k} and \emph{d} have the same number of upper cased characters.         
\item    \texttt{has\_num} : boolean feature that checks if  the strings \emph{k} and \emph{d} have numbers.       
\item \texttt{get\_count} : absolute number of occurrences of \emph{d} in our dataset. 
      \item  \texttt{first\_char\_match} : boolean feature that checks if both \emph{k} and \emph{d} start with the same 5-gram characters.
    \item    \texttt{last\_char\_match} : boolean feature that checks if both \emph{k} and \emph{d} end with the same 5-gram characters. 
    \item    \texttt{w2v\_ad\_sim} : a numeric feature that calculates the cosine similarity between the answer key and distractor using their word2vec  representations.  
    \item      \texttt{wmd\_w2v\_qd} : word mover's distance between the question and distractor using their word2vec vector representations. 
        \item  \texttt{wmd\_w2v\_ad} : word mover's distance between the answer and distractor using their word2vec vector representations. 
    \item     \texttt{glove\_ad\_sim} : the cosine similarity between the answer and distractor using their averaged glove embeddings. 
    \item     \texttt{wmd\_glove\_ad} : the word mover's distance between the answer and distractor using their averaged glove embeddings.
        \item \texttt{lang\_prior} : the prior distribution of the source language of the question.    

\end{enumerate}

\section{Annotation Platform}
\label{appendix:annotationplatform}

\begin{figure}[h!]
\includegraphics[width=\linewidth]{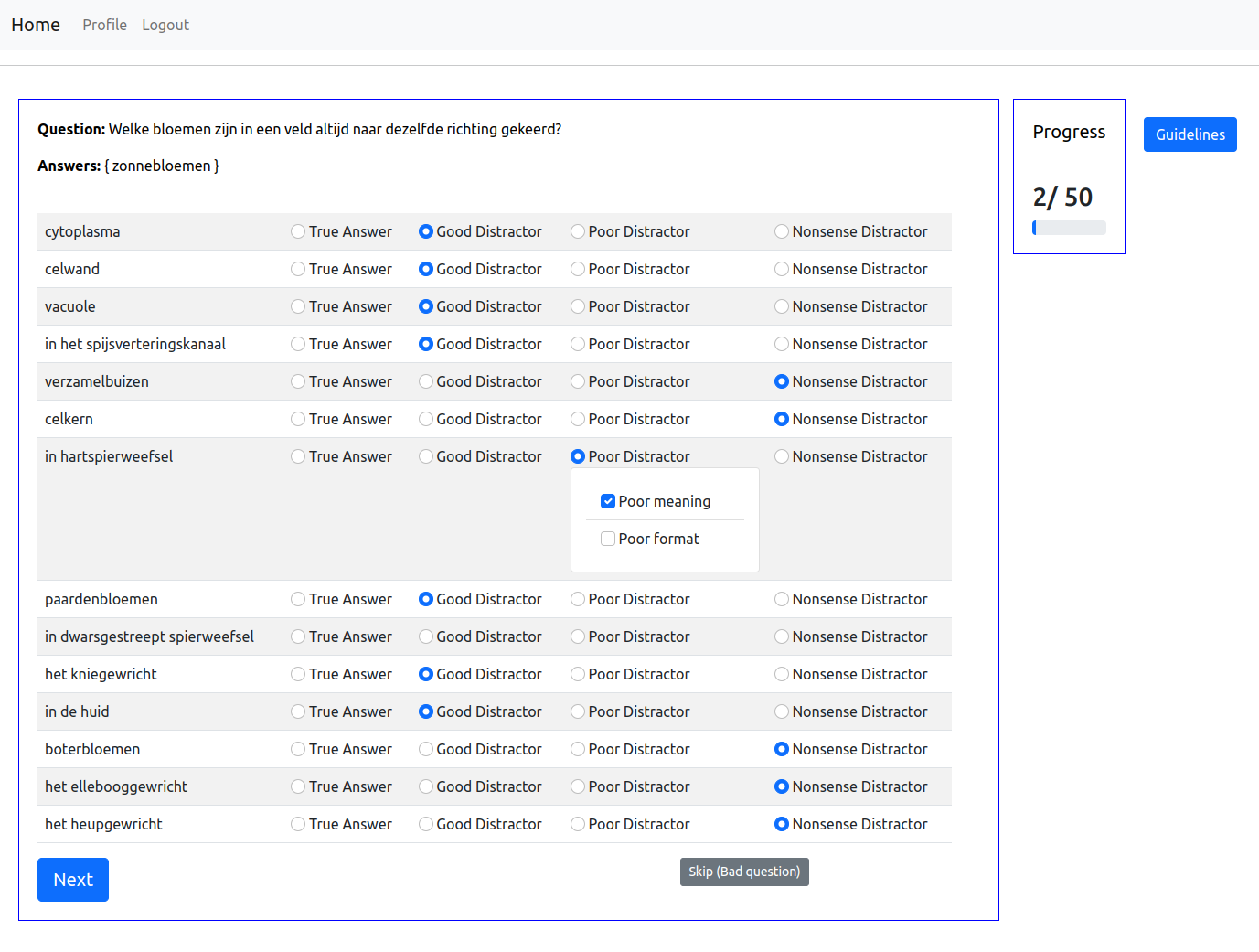}
\caption{Screenshot of the distractor annotation tool. The teacher is shown a question, an answer, and a shuffled list of ground-truth distractors \& candidate distractor suggestions by all the models.}
\label{fig:screenshot}
\end{figure}

\Figref{fig:screenshot} shows the annotation tool that we built for the quality annotation task by teachers. Each page presents a question, its actual answers, and a randomly shuffled list of \rebuttal{candidate distractors that are proposed by the different models. Teachers assign quality labels to each of these proposed distractors by selecting one of the four radio-button options. If the teacher selects \emph{poor distractor} as a label for a distractor,} then 
a drop-down menu with two more options (\ie \emph{poor format} and \emph{poor meaning}) is shown. Finally, \rebuttal{the annotator/teacher} can go to the following question by pressing the `Next' button displayed at the left bottom of the screenshot. 

\section{User study details}
\label{appendix:userstudydetails}

This section contains the user study details. \Tabref{tab:annotationsdatadescription} 
\rebuttal{describes} the data gathered from the annotations provided by the teachers. 
\rebuttal{Every subject has 50 questions except English, which had two duplicates that were later removed, leaving only 48 questions. On average, there are 2 distractors for each question item. We collected 1090 annotations for the original ground-truth distractor, and 11,633 annotations for the proposed candidate distractors (\ie top ten ranked distractors by each of the four models).} 
A total of 8 teachers participated in the study. English \rebuttal{(\ie from languages) and History (\ie from factoids)} were annotated twice by two different teachers \rebuttal{for the purposes of calculating interannotator agreement}.

\begin{table*}[h!]
\centering
\footnotesize
\begin{center}
\caption{Ratings Data Description }
\label{tab:annotationsdatadescription}
\begin{tabular}{lcccccccc}
\toprule
\textbf{Subjects} & 
\multicolumn{1}{c}{\textbf{Item count}} &
\multicolumn{2}{c}{\textbf{Dist. count }} &
\multicolumn{2}{c}{\textbf{Ratings count}} &
\multicolumn{1}{c}{\textbf{No of Raters}} &
\\
& &
\multicolumn{1}{c}{\textbf{Original dist}} &
\multicolumn{1}{c}{\textbf{Proposed dist}} &
\multicolumn{1}{c}{\textbf{Gold dist}} &
\multicolumn{1}{c}{\textbf{Proposed dist}} &
\\\cmidrule(lr){3-4} \cmidrule(lr){5-6}

\\\midrule
English         &  48   & 130   & 723   & 260   & 1882  & 2 \\ 
French          &  50   & 92    & 1148  & 92    & 1650  & 1 \\ 
Geography       &  50   & 145   & 966   & 290   & 1960  & 1 \\ 
History         &  50   & 130   & 1335  & 260   & 2420  & 2 \\ 
Biology         &  50   & 88    & 1266  & 88    & 1761  & 1 \\ 
Nat. Sciences   & 50    & 100   & 1407  & 100   & 1960  & 1 \\
\midrule
Total           & 298   & 685   & 6845  & 1090  & 11633 & 8  \\ 
\bottomrule
\end{tabular}
\end{center}

\end{table*}

\begin{table}[ht]

\begin{center}
\caption{Contingency table for automatic ranking \& human rating correlation using \dqsim{}}
\label{tab:contingency_table_hypo2}
\begin{tabular}{lccc}
\toprule
& 
\multicolumn{1}{c}{\textbf{Plausible}} &

\multicolumn{1}{c}{\textbf{Less plausible}} &

\\\midrule
Ranked top 5    & 425   & 977  \\
Ranked 5--10     & 303   & 1097\\
\bottomrule
\end{tabular}
\end{center}

\end{table}

\Tabref{tab:contingency_table_hypo2} shows the contingency table for hypothesis 2 that tests the correlation between automated distractor rankings (\ie using our best model \dqsim{}) and human ratings using Fisher's exact test. The \emph{plausible} column contains the count of distractors that were rated `good' and the \emph{less plausible} column the count of all distractors that were rated otherwise (\ie `poor', `true answer' and `nonsense' distractors). The rows indicate the count of top-5 ranked distractors and the 5--10 ranked distractors. 

\begin{table}[H]

\begin{center}
\caption{Contingency table for comparing human \& system generated distractors}
\label{tab:contingency_table_hypo3}
\begin{tabular}{lccc}
\toprule
& 
\multicolumn{1}{c}{\textbf{Plausible}} &

\multicolumn{1}{c}{\textbf{Less plausible}} &

\\\midrule
Human-generated     & 511   & 156  \\
System-generated    & 255   & 412\\
\bottomrule
\end{tabular}
\end{center}

\end{table}

\Tabref{tab:contingency_table_hypo3} shows the contingency table for \rebuttal{testing} hypothesis 3 that compares the quality of human-generated with system-generated distractors. \rebuttal{We use Fisher's exact test to test the hypothesis.} The table shows counts of ratings in each category. For the human-generated row, we keep track of how each ground-truth distractor was rated, and update the counts depending on whether the distractors were rated `good' \rebuttal{(\ie \emph{plausible})} or the other labels \rebuttal{(\ie \emph{less plausible})}. Similarly, for the system-generated row, we count the ratings of top-$k$ proposed distractors and update the counts \rebuttal{in each column} accordingly, where k is determined by the number of ground-truth distractors for that specific question.

\begin{table}[H]

\begin{center}
\footnotesize
\caption{Conditional probabilities between raters (average of both directions)}
\label{tab:conditionalprobraters}
\begin{tabular}{lcccccc}
\toprule
\textbf{Sub.} & 
\multicolumn{1}{c}{$gd \mid tr$} &
\multicolumn{1}{c}{$gd \mid pf$} &
\multicolumn{1}{c}{$gd \mid pm$} &
\multicolumn{1}{c}{$gd \mid ns$} &
\multicolumn{1}{c}{$pf \mid ns$} &
\multicolumn{1}{c}{$pm \mid ns$} 

\\\midrule
Eng. & 35\% & 19\% & 44\% & 6\% & 11\% & 14\%   \\
His. & 50\% & 22\% & 34\% & 5\% & 12\% &6\% \\

\bottomrule
\end{tabular}
\end{center}

\end{table}

\tabref{tab:conditionalprobraters} \rebuttal{illustrates} the confusion observed between teachers in choosing which label to assign to a distractor. We show the confusion 
\rebuttal{using} conditional probabilities computed over both directions of the raters, where $gd$, $tr$, $pf$, $pm$, and $ns$ \rebuttal{denote} good, true answer, poor format, poor meaning and nonsense distractors respectively. For example, the first column (\ie $P(gd\mid tr)$ ) shows the probability of rating a distractor `good' given that it was rated `true answer' by the other rater. 

\end{document}